\documentclass[final,5p,times,twocolumn]{elsarticle}

\newcommand\like[1]{\begin{picture}(1,1)
\ifnum0=#1\put(.5,.35){\circle{1}}\else
\ifnum10=#1\put(.5,.35){\circle*{1}}\else
\put(.5,.35){\circle{1}}\put(.5,.35){\circle*{.#1}}
\fi\fi\end{picture}}

\usepackage[utf8]{inputenc}
\usepackage{newunicodechar}
\newunicodechar{ℓ}{l}

\usepackage{lineno}
\usepackage{graphicx}
\usepackage{amssymb}
\usepackage{pdflscape}
\usepackage[english]{babel}
\usepackage[dvipsnames,svgnames,x11names]{xcolor}
\usepackage{booktabs,tabularx}
\usepackage{multirow}
\usepackage{subfig} 
\usepackage{hyperref}
\usepackage{amsmath}
\usepackage{commath}
\usepackage[english]{babel}
\usepackage[ruled]{algorithm2e}
\SetEndCharOfAlgoLine{}
\usepackage[leftmargin=6em,rightmargin=12em,indentfirst=false]{quoting}
\usepackage{siunitx}

\usepackage[final]{changes}
\usepackage{float}

\usepackage{lineno}
\usepackage{tikz}
\usepackage[export]{adjustbox}

\usepackage{graphicx}
\usepackage[utf8]{inputenc}
\usepackage[export]{adjustbox}
\usepackage{wrapfig}
\usepackage{dcolumn}

\usepackage{pgfplots}
\pgfplotsset{width=10cm,compat=1.9}

\usepackage{array}

\newcolumntype{L}[1]{>{\raggedright\let\newline\\\arraybackslash\hspace{0pt}}m{#1}}
\newcolumntype{C}[1]{>{\centering\let\newline\\\arraybackslash\hspace{0pt}}m{#1}}
\newcolumntype{R}[1]{>{\raggedleft\let\newline\\\arraybackslash\hspace{0pt}}m{#1}}

\usepackage{subfiles}

\usepackage{todonotes}

\journal{Building and Environment}
\begin{document}
	
\begin{frontmatter}

\title{Personal thermal comfort models using digital twins: Preference prediction with BIM-extracted spatial-temporal proximity data from Build2Vec}

\author{Mahmoud Abdelrahman, Adrian Chong and Clayton Miller\,$^{*}$}

\address{Department of the Built Environment, National University of Singapore (NUS), Singapore}

\address{$^*$Corresponding Author: clayton@nus.edus.sg, +65 81602452}

\begin{abstract}
Conventional thermal preference prediction in buildings has limitations due to the difficulty in capturing all environmental and personal factors. New model features can improve the ability of a machine learning model to classify a person's thermal preference. The spatial context of a building can provide information to models about the windows, walls, heating and cooling sources, air diffusers, and other factors that create micro-environments that influence thermal comfort. Due to spatial heterogeneity, it is impractical to position sensors at a high enough resolution to capture all conditions. This research aims to build upon an existing vector-based spatial model, called Build2Vec, for predicting spatial-temporal occupants' indoor environmental preferences. Build2Vec utilizes the spatial data from the Building Information Model (BIM) and indoor localization in a real-world setting. This framework uses longitudinal intensive thermal comfort subjective feedback from smart watch-based ecological momentary assessments (EMA). The aggregation of these data is combined into a \emph{graph network structure} (i.e., objects and relations) and used as input for a classification model to predict occupant thermal preference. The results of a test implementation show 14-28\% accuracy improvement over a set of baselines that use conventional thermal preference prediction input variables.
\end{abstract}

\end{frontmatter}


\section{Introduction}
The indoor environment is a crucial factor in human life, health, and well-being as people spend up to 90\% of their lives in buildings \cite{Asadi2017}. Therefore, poor Indoor Environmental Quality (IEQ) impacts productivity \cite{Esfandiari2017InfluenceReview} and physical and mental health \cite{Vilcekova2017}. A basic component of IEQ is the acceptance of thermal comfort parameters of spaces by occupants and their preference for conditions to change or stay the same \cite{kim2012nonlinear, esfandiari2017influence}. Prediction of this dimension traditionally, according to ISO 7730, relies on personal factors such as clothing insulation, metabolic rate ~\cite{Wang2018-aj}, and environmental factors such as indoor temperature, humidity, radiant temperature, and air velocity ~\cite{De_Dear2013-uo,Graham2021-en}. More recent studies include other personal factors such as gender, age, height, weight, acclimatization, and other factors in addition to environmental factors such as lighting, CO2, and noise \cite{Crosby2021-hr}. The pairwise interaction between these factors is an ongoing area of research. Personal and environmental factors have been extensively used in literature to predict occupants' satisfaction through statistical methods and machine learning. A recent prominent analysis of comfort models using only conventional features shows accuracy less than 33\% of the time~\cite{Cheung2019-fs}. This study was conducted using 81,846 complete sets of objective indoor climatic observations with accompanying \emph{right-here-right-now} surveys from dozens of studies worldwide~\cite{Foldvary_Licina2018-sk}. 

To respond to deficiencies in conventional methods, the personal comfort model paradigm was developed recently to collect data more specific to the context of the individual for whom the model is explicitly trying to predict comfort~\cite{Kim2018-tu}. This paradigm formed a foundation in new types of data that were collected for thermal comfort prediction. Some of these innovative data sources include physiological data from wearable devices~\cite{Liu2019-pi,Choi2012-gq}, higher resolution longitudinal data collection and measurements~\cite{Cheung2017-uw,Kim2019-xe}, and thermal cameras~\cite{Aryal2019-bx}. This paradigm has been a catalyst for innovative methods of collecting subjective feedback from building occupants to be created through innovative interfaces such as polling stations~\cite{Lassen2020-xp}, smart watches~\cite{Jayathissa2019-kg}, and QR code-based sustainability stations~\cite{Sood2019-af} and activity-based workspaces~\cite{Sood2020-vz}. Additional factors have also been explored in terms of their correlation to thermal comfort, including carbon dioxide concentrations, noise, and light levels~\cite{Crosby2021-hr,Crosby2019-nf}. Text-based data sources have even been explored for inferring indoor environmental quality of occupants~\cite{Chinazzo2021-fv}.

\subsection{Using spatial data to supplement thermal comfort prediction}
Despite the extensive work in exploring new input features for personal comfort models, the building's spatial data (e.g., windows, doors, furniture, walls, fans, HVAC components, and equipment) and the occupant location in relation to them have not been thoroughly explored. Buildings are all different, and the spaces within them are heterogeneous. For example, airspeed from a ceiling fan or an air conditioning diffuser is known to differentiate significantly over short distances (centimeters). Also, the furniture arrangement in an office space has a significant impact on daylight distribution, which impacts occupants’ thermal, visual, and cognitive perception~\cite{Ko2020}. The \textit{spatial heterogeneity} of buildings makes it challenging to generalize the environmental measurements from one sensor located at one point in the space to the whole space. This problem has been discussed by several studies constituting a gap in modeling IEQ and understanding occupants' satisfaction corresponding to their location~\cite{DeDear2016PervasiveMonitors}. One approach to overcome this issue is the use of continuous monitoring of IEQ parameters. Numerous IEQ sensor networks have been developed as a trade-off to overcome some of the cost and affordability challenges faced by mobile carts and wall-mounted sensors. They can be affordably deployed on a large scale in a building, be located in different spots, and take continuous readings over time \cite{Parkinson2019}. However, the performance and coverage of large-scale IEQ continuous monitoring in buildings remain an issue. They are also limited in their accuracy and require effort to calibrate and maintain \cite{Heinzerling2013,Asadi2017}. Another approach is using wearable sensors to collect real-time data about the occupant near-body environment as they provide insights into the IEQ spatial-temporal variability.  For example, multiple low-cost sensors deployed in a wide building area can capture the indoor environment's variability more than an accurate mobile cart located in one single spot of the building. A recent study by Jayathissa et al. \cite{Jayathissa2020-pv} found that other factors beyond IEQ readings are helpful or even more accurate in predicting thermal comfort preference.

\begin{figure}
    \centering
    \includegraphics[width=0.9\linewidth]{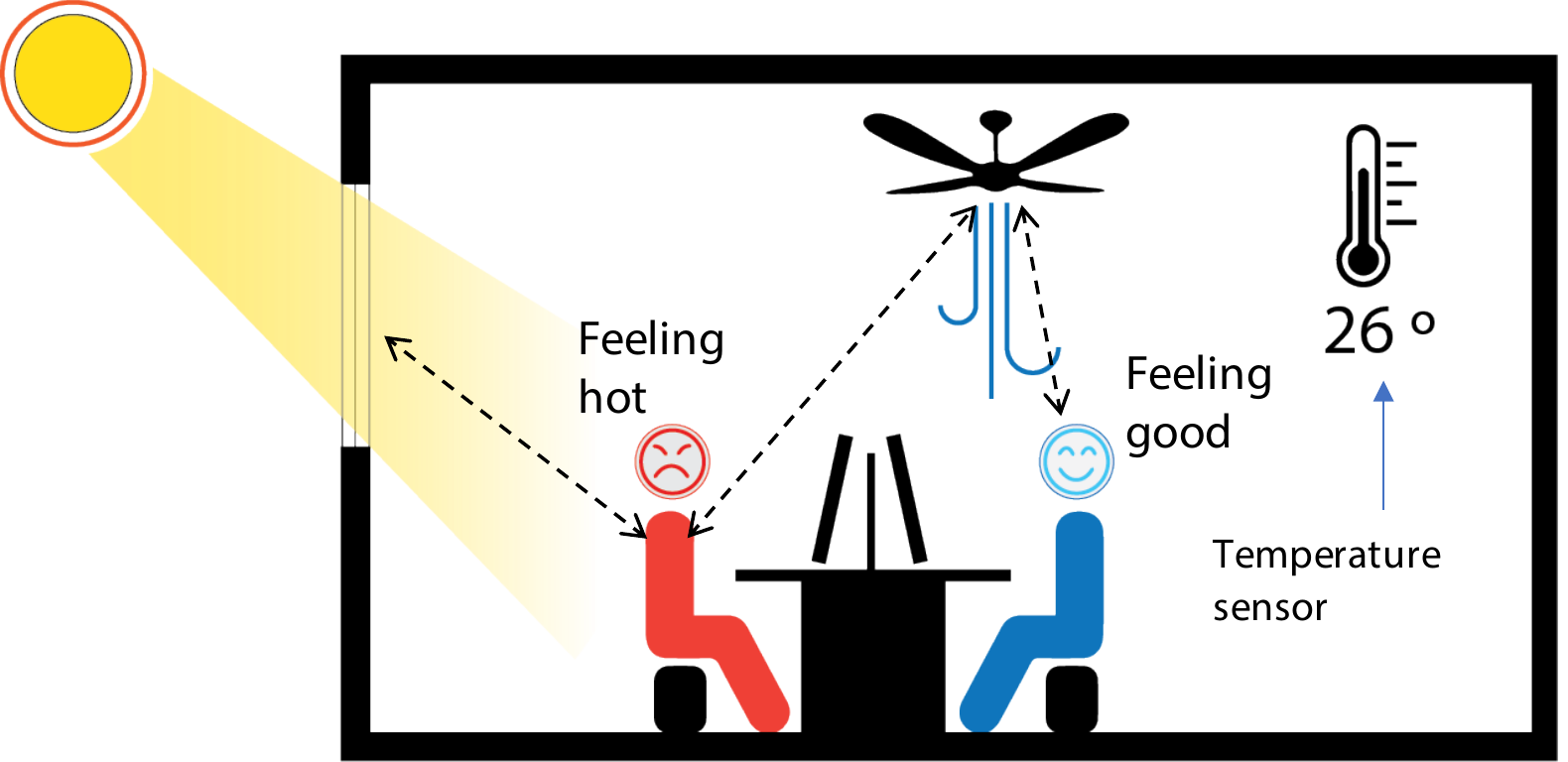}
    \caption{A simple example of occupants in a space that should result in acceptable comfort according to the thermostat. However, the proximity of one user closer to a window and the other closer to an air movement device influences their comfort. Installation of radiant temperature or air movement sensors might capture this discrepancy; however, they are uncommonly used in practice and expensive. Simply using the spatial proximity of these two locations as a feature in the training and testing of personal thermal comfort models could be a proxy for accurate prediction.}
    \label{fig:spatial_proximity}
\end{figure}

The location of an occupant at a certain point in time can influence their overall satisfaction with their environment. A survey conducted on work-space satisfaction showed the importance of variables that are spatially dependent  \cite{Wargocki2012SatisfactionFeatures}. These factors often have a diversity of values within the same space, such as air quality, noise level, visual and sound privacy, and thermal comfort. This variance is attributed to the effect of the space shape and layout, spatial furniture distribution, and proximity to windows, walls, HVAC outlets, or other building objects. Figure \ref{fig:spatial_proximity} illustrates how spatial proximity might impact the thermal comfort satisfaction response of two occupants within the same environment. Traditionally, thermal zones are treated as well-mixed and somewhat homogeneous; therefore, sensors are located in one or more representative locations. These locations are decided according to the pre-defined spatial and temporal resolution based on design guides or codes, regardless of the spatial shape and other heterogeneous aspects of the space. 


\subsubsection{Spatial proximity}
Spatial proximity, or nearness, is a term used in spatial reasoning that refers to the area of impact or influence of each spatial object \cite{Brennan2012SpatialMeasure}. Spatial proximity is not only dependant on distance but also the functionality of the object. For example, a ceiling fan can have different airspeeds and distribution profiles depending on the fan type, size, and the number of blades, as well as the room shape and size \cite{PaulRaftery2020CeilingReport}. Table \ref{tab:spatialProximity} gives an overview of the areas of impact (AoI) of different spatial objects on thermal comfort extracted from literature. Several other spatial proximity factors influence occupants' overall satisfaction, such as visual privacy, comfort, and speech intelligibility. However, this paper only focuses on those related to thermal comfort. 

\begin{table}
  \centering
  \caption{Review on the area of impact (AoI) of different spatial objects on thermal comfort.}
    \begin{tabular}{|p{0.2\linewidth}|p{0.2\linewidth}|p{0.2\linewidth}|p{0.2\linewidth}|}
    \hline
    \textbf{\small{Spatial object}} & \textbf{\small{AoI type}}   & \textbf{\small{AoI size}} & \textbf{\small{Ref}}\\
    \hline
     Ceiling fan & Central & Same as the fan radius at 1.1 m height &\cite{PaulRaftery2020CeilingReport} \\
    \hline
     VAV diffuser & Central& Depends on the grill shape, throw, and spread  & \cite{HuangIntelligent}\\
    \hline
     Window & Linear & 2 - 7 ft from the facade depending on the u-value and the downdraft  & \cite{lyons2000window, huizenga2006window}\\
    \hline
    \end{tabular}%
  \label{tab:spatialProximity}%
\end{table}

\subsection{Previous and related work}
\begin{figure}[hbt!]
    \centering
    \includegraphics[width=\linewidth]{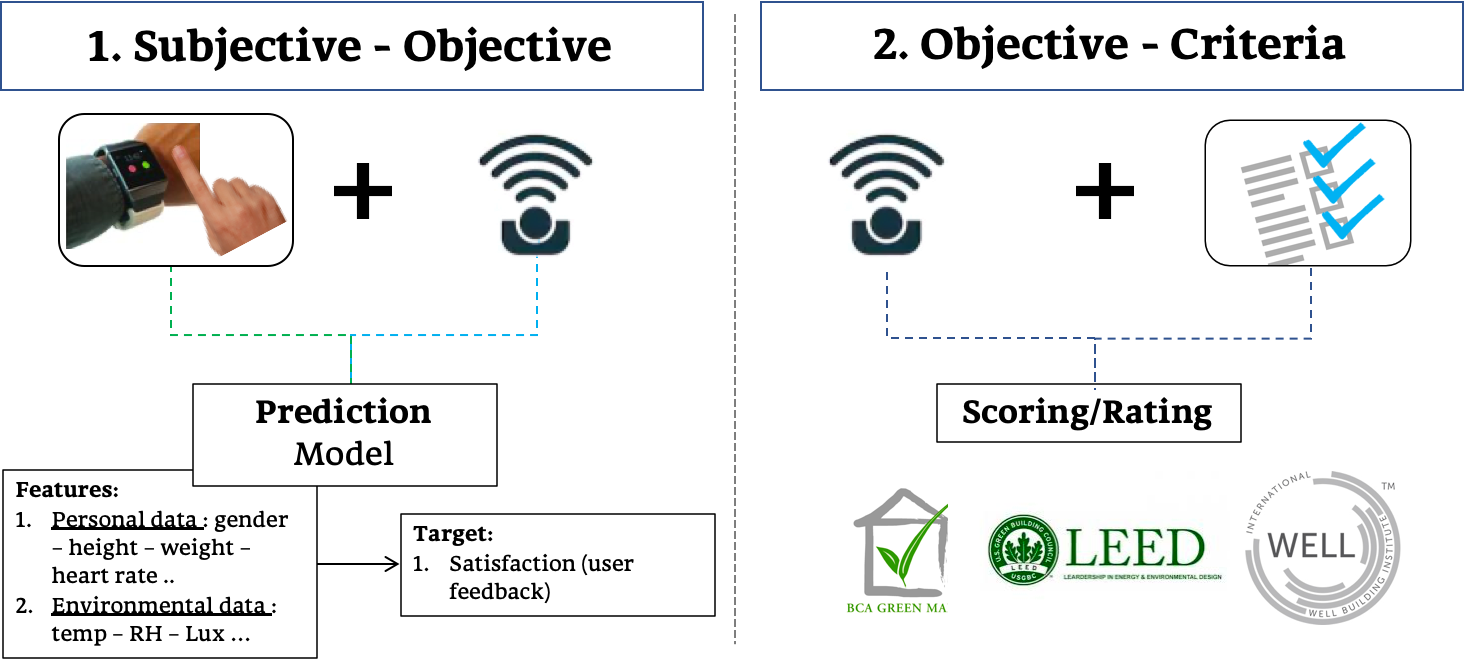}
    \caption{Overview of Subjective-Objective and Objective-Criteria IEQ models}
    \label{fig:ieqmodesl}
\end{figure}

Thanks to the development in low-cost sensor technologies over the past two decades, it has become easier to deploy sensors in buildings to monitor the indoor environment. Using these sensors, building operators have been able to monitor occupant behavior and environmental conditions in the indoor environment in real-time \cite{Ahmad2016BuildingResearch}. This capability supports control strategies tailored to meet occupants' satisfaction either by predicting occupants’ thermal comfort directly from the sensor data \cite{Zou2018} and comparing them with criteria or rating systems (Objective-Criteria). Alternatively, the strategy could include asking people how they feel about the indoor environment (post-occupancy evaluation or POE) and then matching their responses with the sensors to predict their future satisfaction perception using Machine Learning (ML) (Subjective-Objective) \cite{Aryal2019-bx}. Different data science and machine learning models have been used frequently in thermal comfort applications throughout the building lifecycle \cite{Abdelrahman2021-bk}. Luo et al. \cite{Luo2020} conducted a comparative study of different ML models used for thermal comfort prediction. The results showed that Random Forests (RF) and Gradient Boosting Trees (GBT) outperform all the other models, while PMV had the lowest performance.  Figure \ref{fig:ieqmodesl} shows the differences between subjective-objective and the objective-criteria models outlined in the literature.

\emph{Attributed heterogeneous graphs} are data structures used in representing homogeneous static relationships \emph{(homogeneous graphs)} between objects in the form of nodes (also known as vertices) and relations (also known as edges and links). Homogeneous graphs are used to represent simple real-world relations. However, these scenarios are more complex and require richer graphs to define nodes, edges, and all of the whole graph attributes dynamically. Hence, \emph{attributed graphs} are types of \emph{heterogeneous graphs} that have richer data in each node and edge. In the attributed graphs, each node and edge is treated as an object with attributes and functions. For example, if the node represents a building occupant, it will have age, weight, height, and gender attributes. While if the node represents a building's spatial object (e.g., space), it might have attributes such as space name, space type, setpoint temperature,  and the number of occupants. In many cases, nodes can have more complex features such as spatial-temporal data such as location, proximity, and 2D polygon shapes. Also, relations can have spatial-temporal information such as distance or whether an object contains or is a part of another object.

\emph{Graph embedding} is a method generally used to convert a graph (or a part of a graph) into its corresponding vector representation in a low dimensional manifold where the original graph data is preserved. Graph embeddings lie in the intersection between two traditional research fields: 1) representation learning \cite{Bengio2013}, and 2) graph analytics. Most machine learning models work with feature instances in the form of n-dimensional vectors of different types, e.g., numeric data, binary, categorical, as inputs. A successful machine-learning algorithm generally has a quality feature data representation. The motivation behind learning embeddings is to perform essential tasks on networks such as classification \cite{Zhang2016,Chang2015,Perozzi}, prediction \cite{Grover}, and clustering \cite{Tang2016}. For example, predicting the most probable label of a node in a graph could be used as a recommendation system for predicting whether there should be a link between two nodes (friendship recommendation in social networks is a classic example for link prediction).

Building Information Models (BIM) encapsulate building spatial data in a hierarchical graph-like data structure in the built environment domain. \emph{Build2Vec}, developed by Abdelrahman et al. \cite{abdelrahmanbuild2vec} is used to extract the similarities between different spatial objects and locations in a building by embedding its BIM extracted graph into a multi-dimensional vector space. Each spatial object from the BIM model is represented as a vector representing its relation to other objects and its relation to the surrounding environment. 

\subsection{Objectives and novelty}
This paper outlines a framework for extracting spatial proximity information from a BIM model and converges this structure with spatially and temporally dynamic data from a field-based subjective thermal comfort study to create a digital twin modeling environment. The objective is to improve the ability to predict an individual's thermal comfort preference using new spatial-based features extracted from this digital twin context. This study improves the state-of-the-art through the following points:
\begin{enumerate}
    \item While most of the predictive thermal comfort models focus only on the temporal data, there is still a need to investigate the effect of incorporating the location of temporal data and the relation between the occupant and different building objects (spatial objects).
    \item Many of the recent successful studies on thermal comfort prediction have been conducted in real-world scenarios (i.e., not in experimental controlled chambers). However, the majority of which have been restricted to sensing the environment either using static sensors located at a representative position in the space which is mostly constrained by the infrastructure, or using a large number of continuous sensing sensors located on the occupants' workplace to capture spatial heterogeneity. The key question remains about leveraging the spatial objects and spatial-temporal data instead of installing more sensors to capture spatial heterogeneity.
\end{enumerate}

Towards these objectives, Section \ref{sec:methodology} outlines the case study, experimental data set, digital twin data extraction process from the BIM model, and the model testing for thermal preference prediction as compared to a baseline from previous work with the data set. Section \ref{sec:results} showcases the results of the experiment to show the results of spatial similarity calculations and prediction accuracy. Sections \ref{sec:discussion} and \ref{sec:conclusion} discuss the impact these results could have on the increased use of BIM models and digital twins in the characterization of occupant thermal comfort characterization.

\section{Methodology}
\label{sec:methodology}

This work builds upon a previous study with 30 participants involved in an experiment that collected dynamic preference data from occupants about their thermal, lighting, and noise-based preferences \cite{Jayathissa2020-pv}. This experiment was deployed using the Cozie application, which is an open-source, community-driven human-building interface for subjective occupant feedback built on the Fitbit and Apple Watch platforms\footnote{\url{https://github.com/cozie-app}}. This study also collected temporal data from fixed IEQ sensors in the indoor environment (temperature, humidity, noise, and carbon dioxide) and physiological data from the smartwatch (heart rate, near body temperature). The data from this study was used to build personal comfort models~\cite{Jayathissa2020-pv}, analyze the comfort behavior for the transition period between spaces~\cite{Sae-Zhang2020-fy}, and understand the value of class balancing for thermal comfort prediction~\cite{Quintana2020-zu}. This paper builds upon this work by adding the building's spatial data (from the BIM model) and the spatial-temporal data of occupants related to their environment (location of people as compared to the sensors). This section outlines the methods for testing the addition of spatial objects and the spatial-temporal data into the prediction models for comparison against the existing personal models as baselines. 


\subsection{Case study}
\begin{figure*}[!h]
    \centering
    \includegraphics[width=\linewidth]{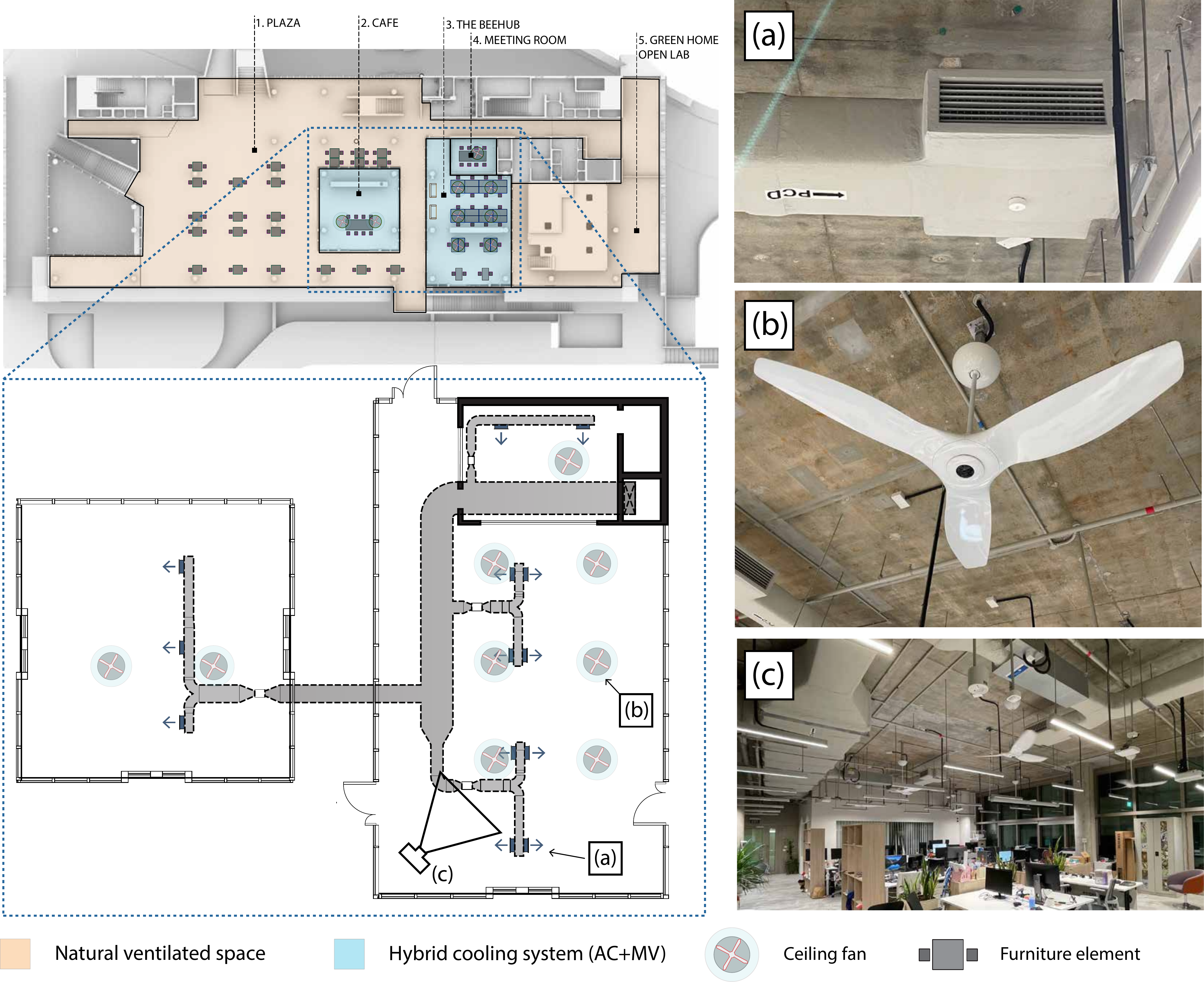}
    \caption{An illustration of the floor plan for this case study located in the third floor of SDE4. The floor consists of five main spaces with two of them being naturally ventilated, and three spaces are hybrid cooled.}
    \label{fig:illustration_of_floor3}
\end{figure*}

The selected case study, SDE4, is an educational building located at the National University of Singapore (NUS). This Net-Zero Energy Building (NZEB) is selected as a pilot case study for the experiment as it is a test-bed for numerous digital and sensor-based technologies. With a gross floor area of 8,517 square meters divided into six floors, SDE4 has a mixture of enclosed and open spaces. The building consists of 86 zones with functions such as laboratories, teaching classrooms, design studios, a library, administrative offices, mechanical systems rooms, storerooms, and shafts. Many of the zones are naturally ventilated (NV), including transitional (e.g., corridors), semi-open spaces (e.g., plaza and terraces), and service rooms (e.g., toilets and storage). Other zones are either mechanically ventilated (MV) or include traditional air-conditioning (AC). Many spaces adopt a hybrid cooling system (HC) that employs both AC and ceiling fans with a design zone setpoint temperature of 27 degrees Celsius and elevated air movement via ceiling fans. Ceiling fans provided thermal comfort with higher setpoint temperatures based on predicted mean vote (PMV) calculations and validated experiments \cite{Qi2017}. The current study only focuses on the third floor of SDE4, as shown in Figure \ref{fig:illustration_of_floor3}. 

\begin{figure*}
    \centering
    \includegraphics[width=\linewidth]{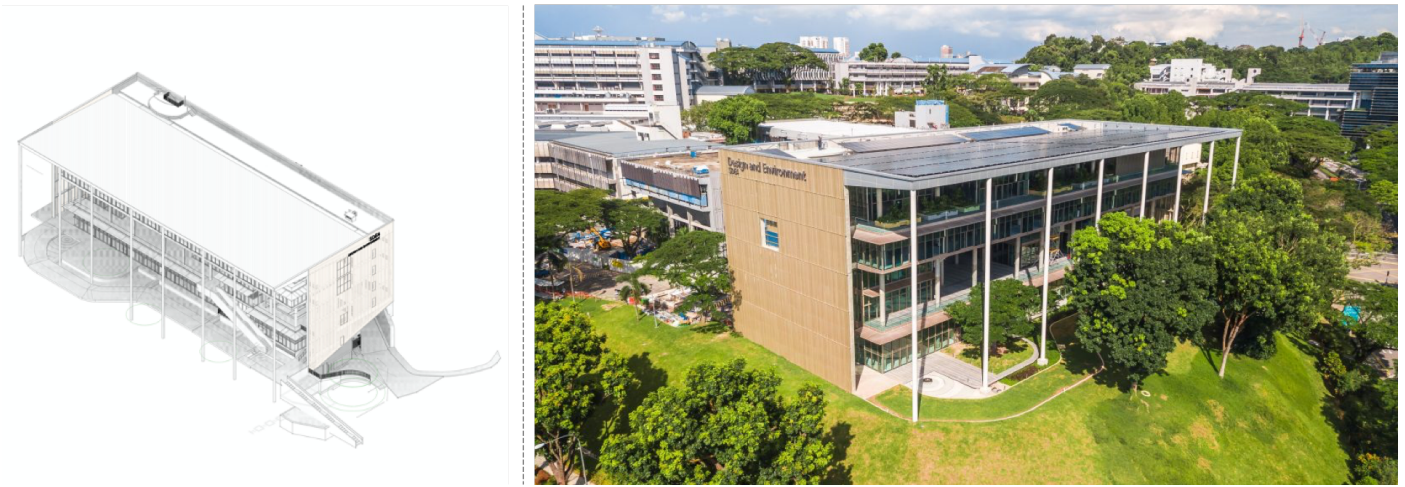}
    \caption{The SDE4 building used in the case-study. The photo on the right is adapted from the NUS Office of Estate Development (OED) \cite{SDEOED}.}
    \label{fig:case_study_3d}
\end{figure*}


\subsection{Data collection and processing}

Based on the previous experiments and the proposed methodology, there are four main types of data used in this research, which are explained in this subsection:
\begin{enumerate}
    \item Environmental data from sensors (Objective-temporal). 
    \item Occupant's satisfaction feedback (Subjective-temporal).
    \item Occupant's location in the building (Spatial-temporal)
    \item Building spatial data and spatial representations
\end{enumerate}

\begin{figure*}[!h]
    \centering
    \includegraphics[width=\linewidth]{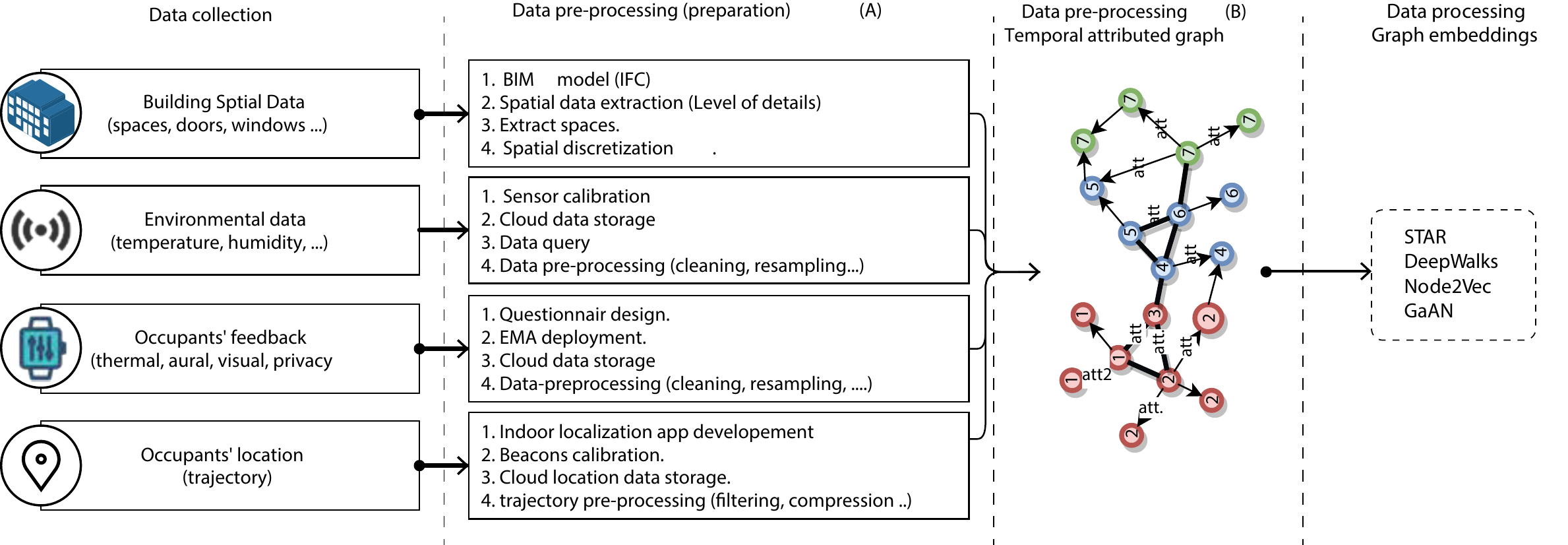}
    \caption{Summary of the data collection, pre-processing, and processing for the Build2Vec model}
    \label{fig:data_collection_plan}
\end{figure*}

\begin{figure}[!h]
    \centering
    \includegraphics[width=\linewidth]{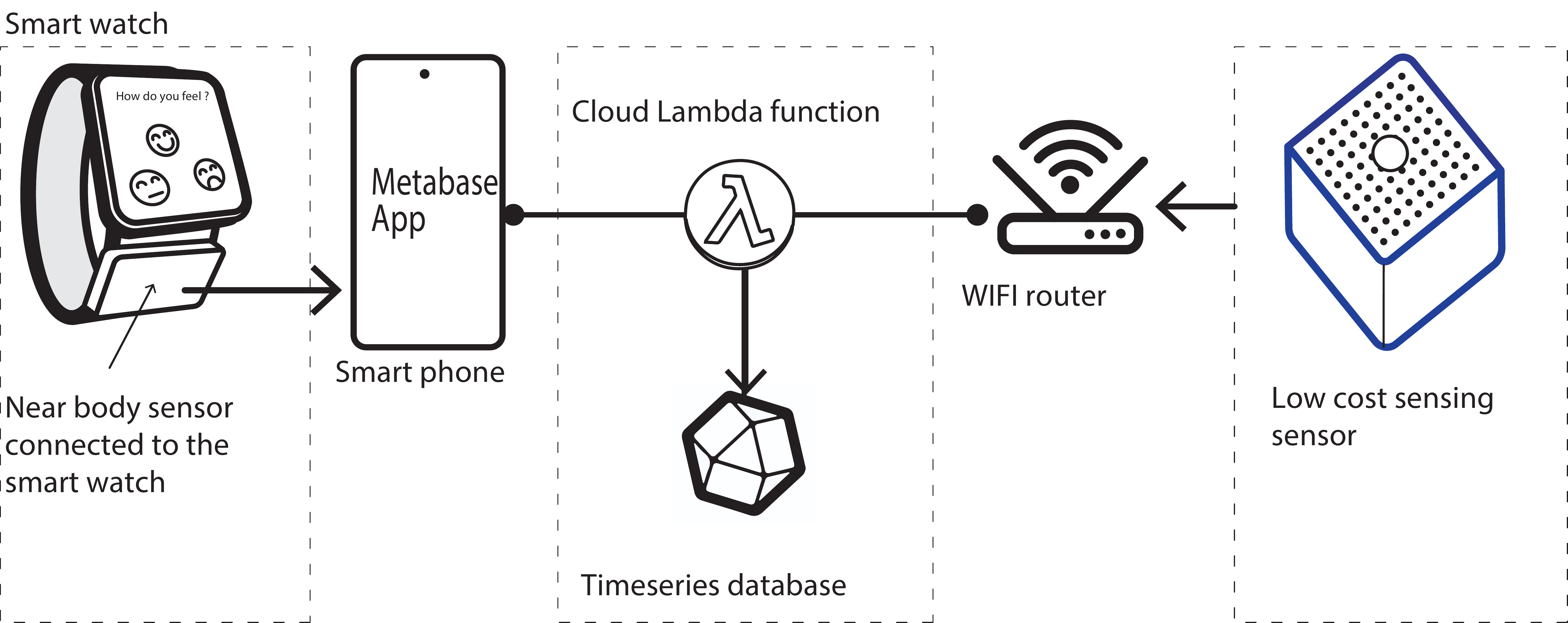}
    \caption{Environmental and occupant data collection pipeline overview for the experiment.}
    \label{fig:data_collcetion_infrastructure}
\end{figure}

The overview structure of these data sources and how they are used are summarized in Figure \ref{fig:data_collection_plan}. For the first three data sources, there is an extensive amount of explanation in previous studies using this data set \cite{Jayathissa2020-pv}. In order to get a sense of accuracy ranges and technologies, a list of the sensors used in the experiment and their specifications is shown in Table \ref{tab:sensors_used}. Figure \ref{fig:data_collcetion_infrastructure} shows the hardware and software infrastructure used to collect the data from occupants and the environment, including the data pipeline flow.

\begin{table*}[!h]
\centering

  \caption{Sensors manufacturer and measurement specifications from the experimental deployment in SDE4.}
\label{tab:sensors_used}
\begin{tabular}{@{}llll@{}}
\toprule
Equipment                        & Parameters                          & Measurement                                                                                    & Accuracy              \\ \midrule
\multirow{2}{*}{Fitbit Versa2}   & \begin{tabular}[c]{@{}l@{}}Thermal comfort \\subjective feedback\end{tabular} & \begin{tabular}[c]{@{}l@{}}Prefer warmer,\\ Comfy, \\ Prefer cooler\end{tabular}               & \multicolumn{1}{c}{-} \\ \cmidrule(l){2-4}
                                 & Heart rate                          & Beats per munite                                                                               & 2.5 bpm               \\\midrule
mBient MetamotionR               & Near-body temperature               & \textbf{\textdegree C}                                                                                    & 0.5 \textdegree C                \\\midrule
SteerPath                        & Indoor localization                 & \begin{tabular}[c]{@{}l@{}}Longitude,\\ Latitude,\\ Accuracy,\\ Floor,\\ Space id\end{tabular} & 0.25 to 5m            \\\midrule
\multirow{4}{*}{SENSing sensors} & Air temperature                     & \textdegree C                                                                                             & 0.5-2.0  \textdegree C            \\\cmidrule(l){2-4}
                                 & Relative humidity                   & \%                                                                                             & 3\%                   \\\cmidrule(l){2-4}
                                 & Noise level                         & dB                                                                                             & 2dB                   \\\cmidrule(l){2-4}
                                 & Illuminance                         & lux                                                                                            & 25lux                 \\ \cmidrule(l){1-4} 
\end{tabular}
\end{table*}

Spatial data are collected from the BIM model using Industry Foundation Classes (IFC) file. 
To reduce the computational burden, only useful spatial objects (generally known as \texttt{IfcObject} were extracted from the BIM model. For example, spaces (\texttt{IfcSpace}\{\texttt{IfcSpace}), doors (\texttt{IfcDoor}), windows (\texttt{IfcWindow}), furniture (\texttt{IfcFurniture}), and sensors (\texttt{IfcSensor}). 
Elements that are contained in a space such as furniture, and distribution elements can be accessed in the \texttt{IFC} using the objectified relationship\\ \texttt{IfcRelContainedInSpatialStructure}. Each tangible object has a spatial (geometric) representation by which the spatial data of the object can be extracted. For example, each space has a location placement (as a list of coordinates \texttt{[x, y, z]}) that is related to the building story \texttt{IfcBuildingStorey}. Additionally, the footprint of the space (the space geofencing or boundary) is also defined in IFC in a number of entities such as: \texttt{IfcBoundedCurve}, \texttt{IfcPolyline}, and \texttt{IfcTrimmedCurve}. For complex shapes the \texttt{IfcGeometricCurveSet} object is used. The 3D geometry of the building can be represented using \texttt{IfcExtrudedAreaSolid}, \texttt{IfcArbitraryClosedProfileDef} and \texttt{IfcArbitrary- ProfileDefWithVoids}. These entities are used to extract each space's boundary as well as different spatial objects (e.g., furniture, windows, doors, and walls). However, there were two main issues during this process. The first challenge is that the BIM model's initial design is not updated with the ground-truth settings. For example, furniture type and location change over time. Secondly, the BIM model follows a local coordinate system to define the spatial representations of each entity. However, the occupant spatial data follows the global Geographic Information System (GIS) coordinate system. Thus switching between coordinates has to be done manually.

\subsubsection{Spatial dicretization and graph formation}
Spatial discretization is used to divide each space into cells of nodes. Each cell has a location (x, y) and identification (e.g., C3010001, C3010002). Whenever a cell falls within a spatial object's area of influence (AoI), this cell is connected to the spatial object's node in the graph structure. For example, the AoI of a ceiling fan is known to be the same as the fan radius, and it fades away as the distance outwards increases \cite{raftery2020ceiling}. Thus, any cell that falls within this AoI will have its node connected to the ceiling fan node.

\begin{figure}
    \centering
    \includegraphics[width=0.8\linewidth]{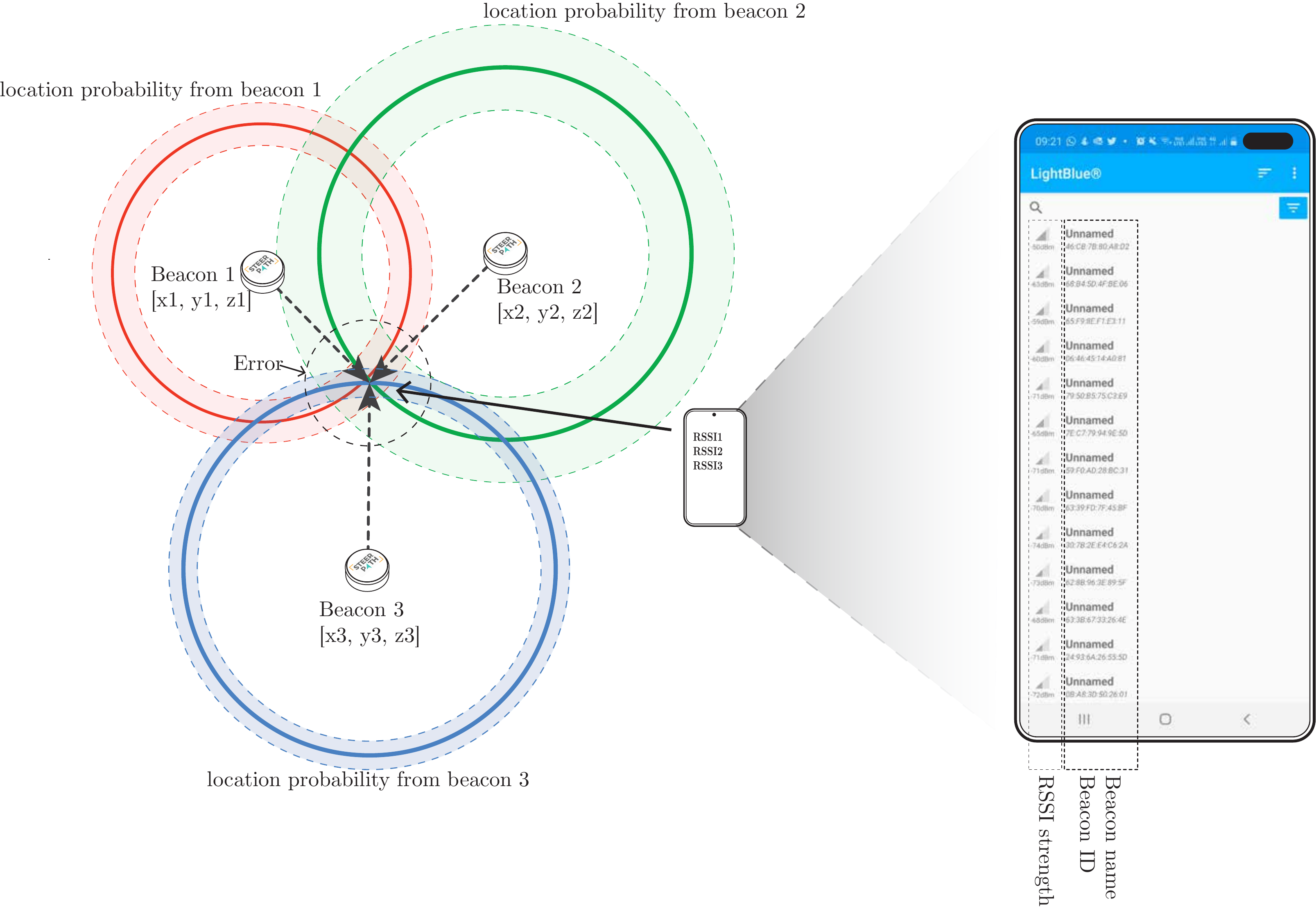}
    \caption{BLE triangulation method, the location of the smartphone falls in the intersection between the three beacons' ranges. The error comes from the obstacles, the attenuation, and sometimes weather conditions.}
    \label{fig:ble_triangulationAsset}
\end{figure}

\begin{figure}
    \centering
    \includegraphics[width=0.8\linewidth]{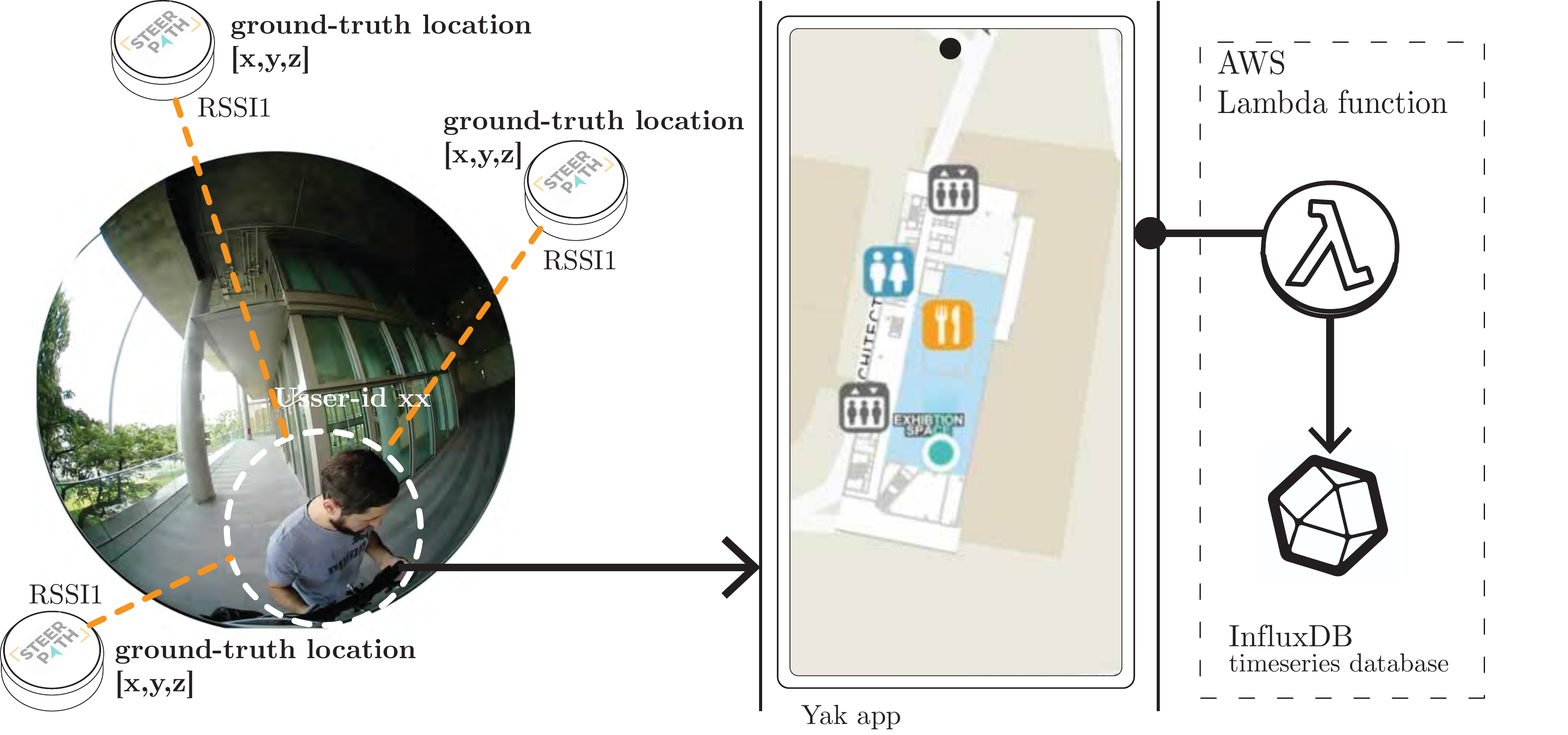}
    \caption{The indoor localization process in which Bluetooth beacons are deployed in the building and participants use an indoor localization app on their smartphone. When an occupant moves in the building, their location is sent to a time-series database that can be written to and read from in real-time.}
    \label{fig:indoor_location_flowAsset}
\end{figure}

The occupants' indoor location is collected using Bluetooth Low Energy (BLE) beacons. Figure \ref{fig:ble_triangulationAsset} illustrates how the global coordinate of each user is calculated using the triangulation method by utilizing the Received Signal Strength Indicator (RSSI) from three different locations simultaneously. The location data is collected in real-time and sent to a time-series database using a smartphone app called YAK \cite{YAK:PDF}. Figure \ref{fig:indoor_location_flowAsset} illustrates how these data are collected and stored during the experiment. This platform collects data that includes \emph{user\_id}, \emph{longitude}, \emph{latitude}, \emph{elevation}, \emph{floor level number}, \emph{timestamp}, and \emph{accuracy}.

Usually, there are two main issues while collecting indoor location data: 1) noise (errors) caused by the inaccuracy of the beacons or the existence of obstacles, and 2) the massive amount of unnecessary data points collected many times. For example, if the temporal resolution is one second, this will result in many data points (up to millions) per user per day, which is not practical. Additionally, not all these data points are helpful. Thus, a pre-processing layer is implemented online (i.e., once the data is generated). 

\subsection{Creation of attributed graphs from BIM model}
\label{sec:attributed_graph}

\begin{figure*}
    \centering
    \includegraphics[width=0.9\linewidth]{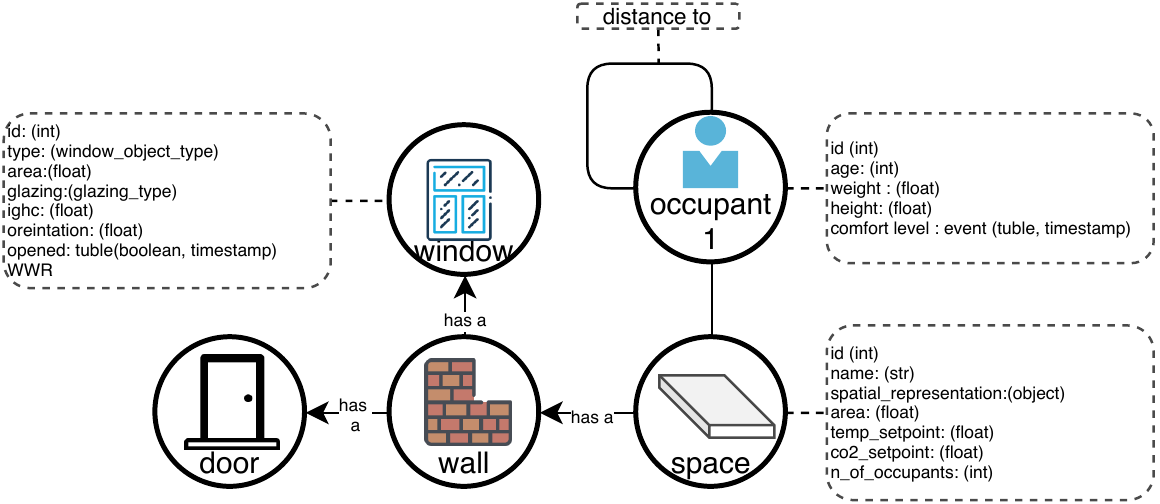}
    \caption{An illustrative example of attributed graphs (heterogeneous graphs), where each node and edge can have one or more types of attributes including timestamp (for temporal attributed graphs).}
    \label{fig:attributed_graphs_example}
\end{figure*}

\begin{figure*}
    \centering
    \includegraphics[width=0.7\linewidth]{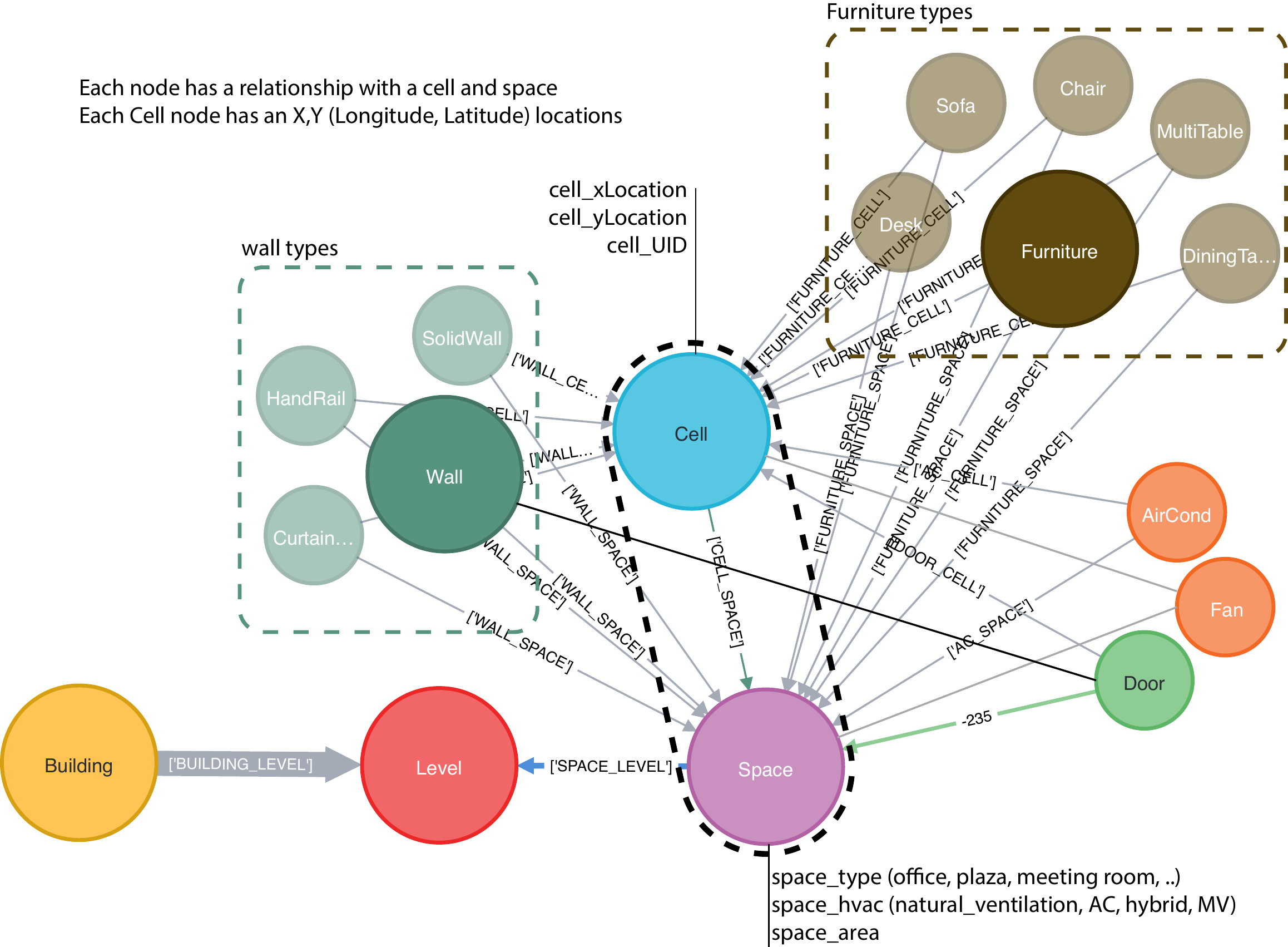}
    \caption{Graph schema visualization where each spatial element is connected to both space and cell nodes.}
    \label{fig:graphSchema}
\end{figure*}

The BIM spatial data are inherently arranged in a graph structure while the occupant dynamic location and data are continuous in space and time. Therefore, it is important to fit the occupant's data into that graph structure. Also, the thermal comfort preference feedback points used in the training model need to be merged into this graph. Hence, we use the K-Nearest Neighbor algorithm to link the occupants' location and thermal comfort votes to the BIM graph model based on proximity. Figure \ref{fig:attributed_graphs_example} shows a schematic example the graph schema of this process and \ref{fig:graphSchema} illustrates a more complex relationship scenario that incorporates a broader range of elements. Table \ref{tab:graph_values} shows the number of each of the elements as extracted from the case study.

\begin{table}[]
    \centering
    \caption{The count of each node label in the graph structure.}
    \label{tab:graph_values}
     \begin{tabular}{lr}
\toprule
{} &  \_labels \\
\midrule
:Cell                  &     4647 \\
:Chair:Furniture       &       86 \\
:DiningTable:Furniture &       24 \\
:Desk:Furniture        &       20 \\
:SolidWall:Wall        &       18 \\
:AirCond               &       15 \\
:HandRail:Wall         &       13 \\
:CurtainWall:Wall      &       11 \\
:ThermalComfortPersonality        &       10 \\
:Fan                   &        9 \\
:Door                  &        7 \\
:Space                 &        5 \\
:Furniture:MultiTable  &        3 \\
:Furniture:Sofa        &        2 \\
:Level                 &        1 \\
\bottomrule
\end{tabular}
    
\end{table}

\begin{figure*}[!h]
    \centering
    \includegraphics[width=\linewidth]{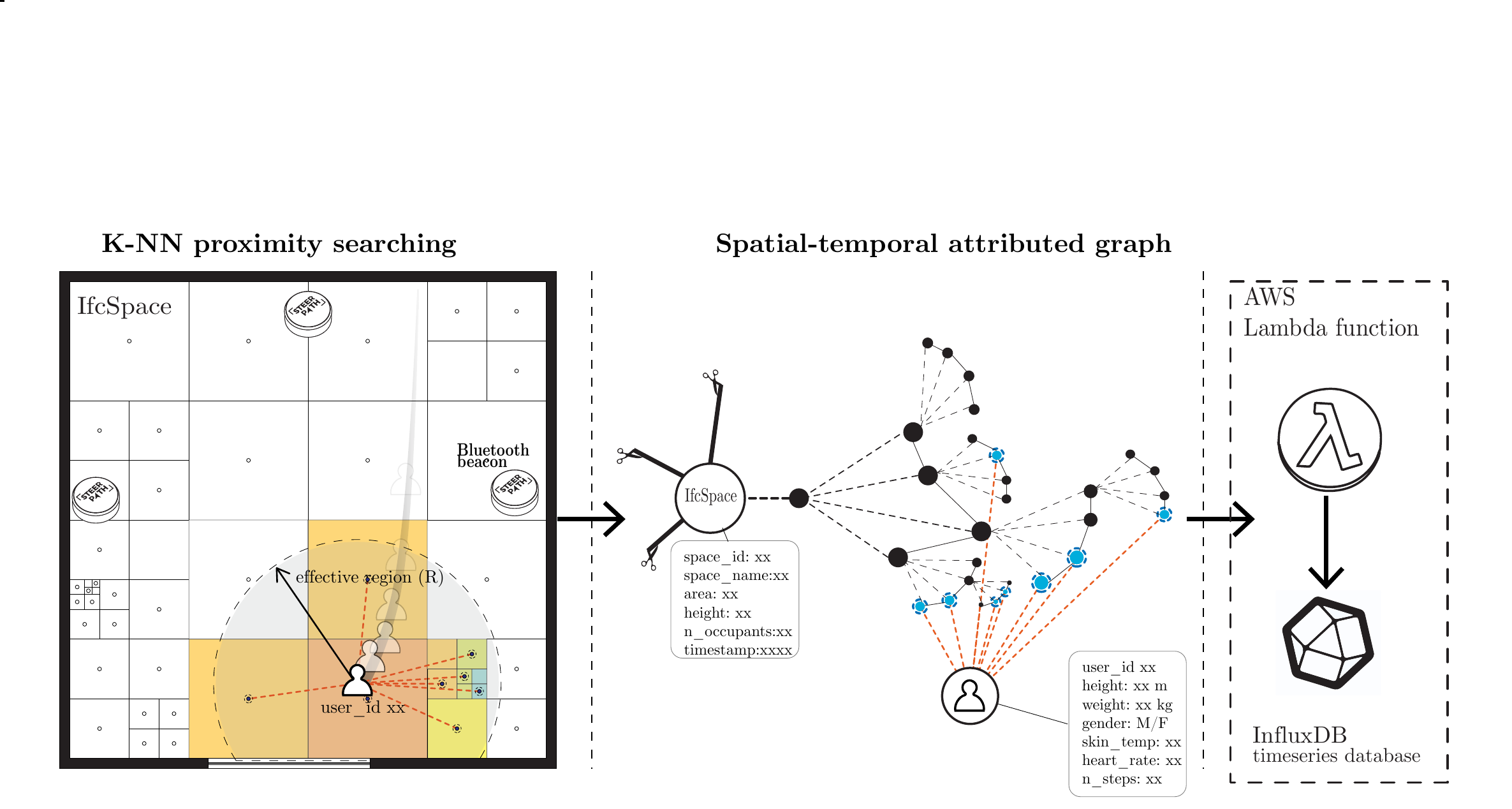}
    \caption{HNSW - K-Nearest Neighbour searching algorithm to extract proximity of the occupant to different spatial object.}
    \label{fig:KNNAsset1}
\end{figure*}

Proximity is used to form the graph structure. Since the [x,y] locations of each spatial cell element are known, and the [x,y] of the occupant is captured in real-time. Then, a K-Nearest Neighbor (KNN) searching algorithm is used to search for the nearest objects. The K-nearest neighbor uses a Euclidean distance range to find the nearest objects within this range. The algorithm must be efficient, given a large number of meshing elements. The na\"ive is based on searching for all the nodes and selecting those with the shortest distances. However, this process is highly complex and scales linearly with the number of nodes, making it infeasible given the large dataset size. Figure \ref{fig:KNNAsset1} shows the state-of-the-art Hierarchical Navigable Small World (HNSW) \cite{Malkov} algorithm that was used to propagate the proximity data to each spatial cell/object~\cite{Wang2013}.

\subsection{Attributed graph embeddings and prediction models}
Attributed graphs are those that have more auxiliary information for both nodes, edges, or the whole graphs. These auxiliary information could be \emph{categorical (labels)}, \emph{numerical (attributes)}, \emph{spatio-temporal}, \emph{node-feature}, \emph{information propagation}, or \emph{knowledge-based}. The prediction operations on different types of attributed graphs generally aim to produce one of the following four types of outputs: node embeddings, edge embeddings, part-of-graph embeddings, and whole-graph embeddings. An object embedding is a vector representation of that object in lower-dimensional space. For example, a graph structure of 10,000 nodes can be represented as an adjacency matrix of 10,000 x 10,000, which is computationally inefficient. Thus, node embeddings of this graph can represent the same data in way fewer dimensions, e.g. (10,000 x 20) where each row is a node, and each column is an embedded feature. Each feature can store different node attributes, such as but not limited to its topography, degree, or betweenness centrality. 

In this research, a vector representation of building data called Build2Vec is used \cite{abdelrahmanbuild2vec}. This model is used to assign an embedding vector to each Node based on their similarity. For example, a node that represents a hybrid cooling space cell located below a ceiling fan will be similar to other nodes located below ceiling fans of the same settings. The power of this model is that it can capture more complex relations automatically from the spatial arrangements. We use feedback from occupants in their responses of thermal comfort preference as a label for that spatial location to be used for training and testing. This concept could be applied to other indoor environmental satisfaction problems such as light, noise, furniture arrangement, amount of space, and privacy. 

\subsubsection{Build2Vec thermal comfort preference model setup}
The Build2Vec model is formed using spatial proximity and the K-NN algorithm. Then, node attributes have been extracted as separate nodes. For example, a node named \emph{natural-ventilated space} was created and connected to all the NV spaces. After that, we made an adjacency list that represents the connection between each pair of nodes. This adjacency list constituted the input to the Build2Vec model. The output from the Build2Vec model is an embedding vector that represents each node based on how similar nodes are to each other.

The embedding vector of each cell node is used as a feature input to the ML classification problem. We used \textit{Random Forest (RF)} classifier from the Python Scikit-Learn package to classify the thermal comfort label of each cell. For example, if a user is feeling a particular thermal preference at a specific cell location, then the feature vector of this cell will be the embedding vector, and the class of the cell based on occupant subjective feedback (\emph{prefer cooler}, \emph{no preference}, or \emph{prefer warmer}). We use a labeled dataset from a previous experiment \cite{Jayathissa2020-pv} where each thermal comfort location and value are known. We use this dataset for training, testing, validation, and benchmarking to the baseline models. K-fold cross-validation is used to validate the models with 30 folds. Each fold consists of 3\% test to train ratio, and the dataset was shuffled for better sampling. The hyperparameters of the RF model are as follows: 
\begin{itemize}
    \item \textbf{Number of estimators} (i.e., the number of trees in the forest) is 200.
    \item \textbf{Max Features} (i.e., the size of the random subsets of features to consider when splitting a node) is set to the default value, which is the square root of the total number of features (total of 22 features -- 20 features from the embedding vector besides heart rate and near-body temperature). 
    \item \textbf{Max depth} (i.e., the maximum depth of the tree) is set to 220.
\end{itemize}



\begin{figure*}
    \centering
    \includegraphics[width=\linewidth]{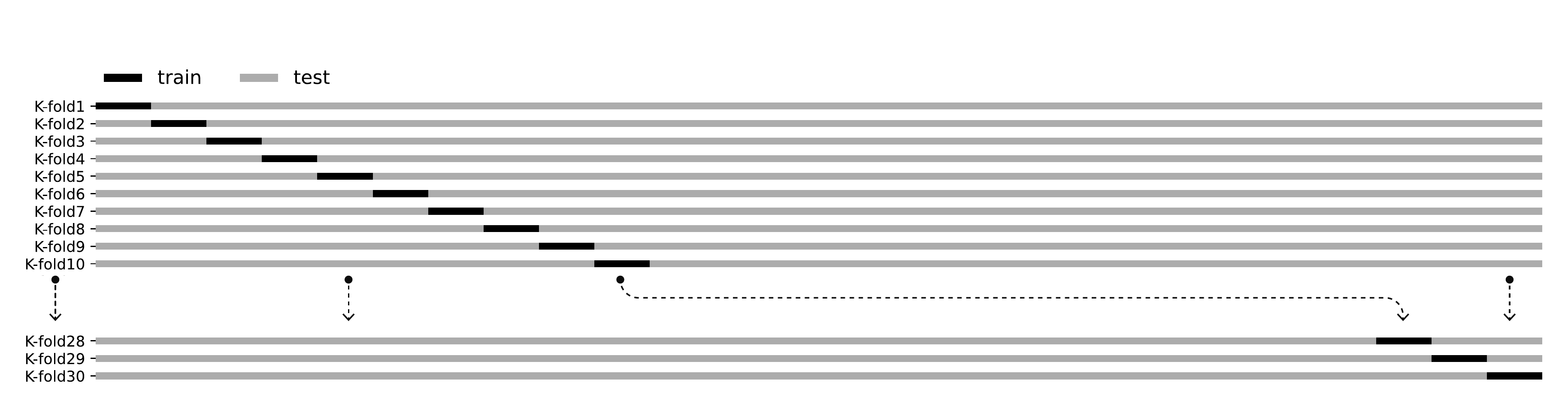}
    \caption{K-fold cross validation using 30 folds was used in the ML process.}
    \label{fig:graphdb}
\end{figure*}

\section{Results}
\label{sec:results}
In this section, the output of applying the methodology is shown on the data from the SDE4 case study. A spatial-temporal graph embeddings algorithm, Build2Vec, was used to classify each user's preference (\emph{prefer warmer, no preference, prefer cooler}) based on the attributes of the user, as well as the spatial and temporal characteristics of the building and the environment. These results enable the analysis of zones that can be considered as having similar comfort preferences based on the model. In addition, the proposed modeling method can be compared to the baseline models from previous work.

\subsection{Build2Vec similarity between cells}
\begin{figure*}[!h]
    \centering
    \includegraphics[width=\linewidth]{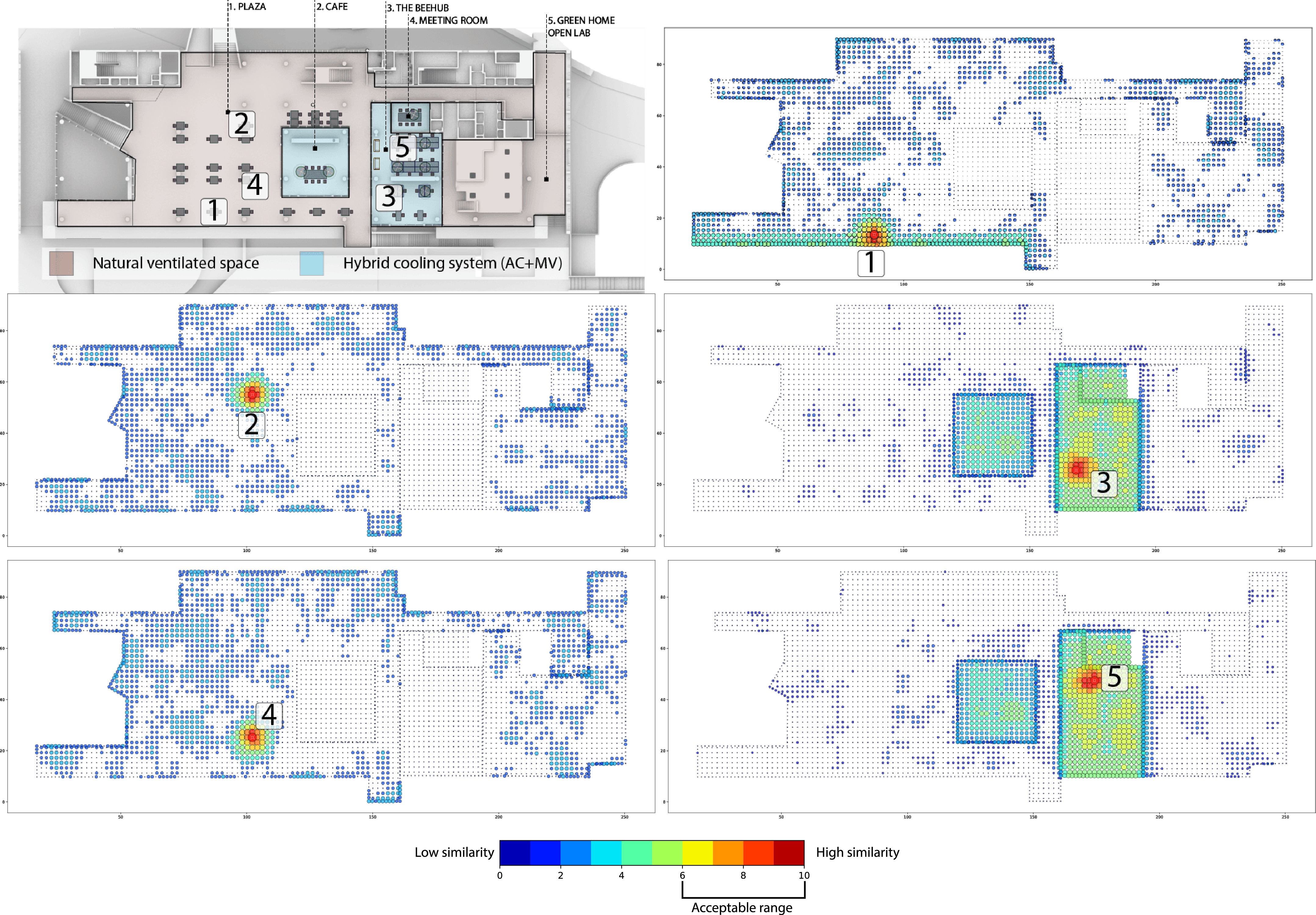}
    \caption{The thermal comfort spatial similarity between cells after applying 
    Build2Vec for five different example points on the floor plan. This spatial similarity is used as an input to the Build2Vec-enhanced preference prediction model.}
    \label{fig:spatial_similarity}
\end{figure*}

\begin{figure*}[!h]
    \centering
    \includegraphics[width=\linewidth]{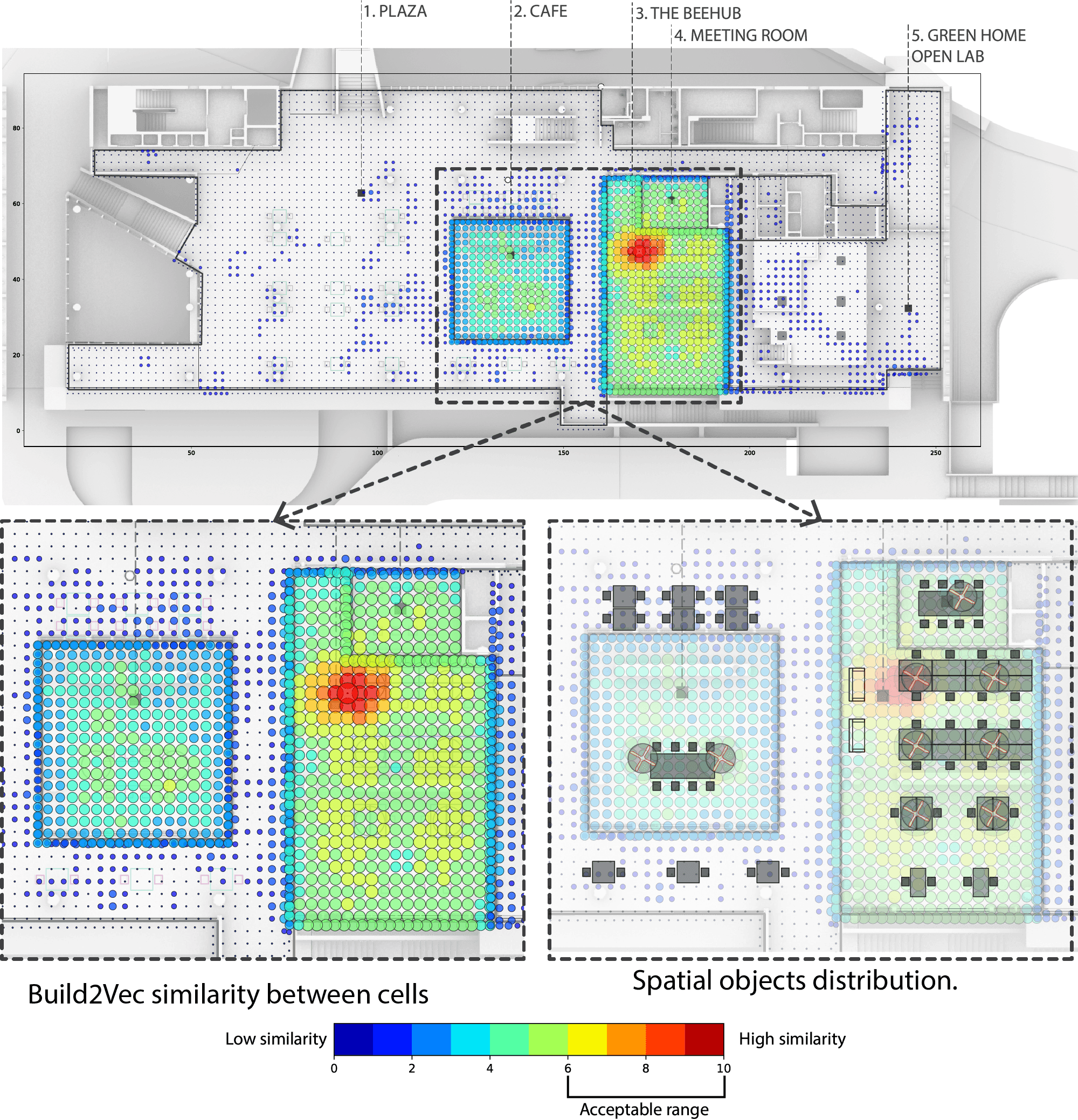}
    \caption{The thermal comfort spatial similarity between cells after applying Build2Vec. The similarity is at its highest values at the proximity of the test points, a fact that can be attributed to the auto-correlative nature of spatial-temporal data.}
    \label{fig:spatial_similarity_zoomin}
\end{figure*}

After applying the Build2Vec model to the spatial structure, each cell has been assigned to an embedding vector. We use the cosine similarity to calculate the similarity between each two cell vectors of an inner product space using Equation \ref{eqn:sim}.

\begin{equation}
\text { similarity }=\cos (\theta)=\frac{\mathbf{A} \cdot \mathbf{B}}{\|\mathbf{A}\|\|\mathbf{B}\|}=\frac{\sum_{i=1}^{n} A_{i} B_{i}}{\sqrt{\sum_{i=1}^{n} A_{i}^{2}} \sqrt{\sum_{i=1}^{n} B_{i}^{2}}}
\label{eqn:sim}
\end{equation}
where $A_{i}$ and $B_{i}$ are components of vector $A$ and $B$ respectively. Figures \ref{fig:spatial_similarity} and \ref{fig:spatial_similarity_zoomin} show how each cell is similar to the other cells in the floor plan using this similarity calcuation.

\subsection{Prediction accuracy}
The baseline model used in this experiment is a thermal comfort right-here-right-now survey of 30 participants conducted in SDE4 \cite{Jayathissa2020-pv}. Each participant was asked to fill in an initial onboarding survey that included physiological and psychological questions related to gender, height, weight, and acclimatization to the local climate. These data were used to cluster users into groups of similar preferences. These clusters were used to predict the thermal comfort using a Random Forest classification of thermal preference with a temporal feature set of data that included timestamps, environmental sensors, near-body temperature, heart rate, room identification, and historical preference. Several models were built using subsets of these features as well. Table \ref{table:results} shows the results of these baseline models.


\begin{table*}[]
\centering
\caption{Results of applying Build2Vec to thermal preference versus the baselines from previous work~\cite{Jayathissa2020-pv}.}
\label{table:results}
\begin{tabular}{@{}l|c|c|c|c|c|c|c@{}}
\toprule
\multicolumn{1}{c|}{\textbf{Feature Sets}} & \multicolumn{6}{c|}{\textbf{Baseline experiments}}  & \textbf{Build2Vec}  \\ \midrule
\textbf{User location}                 &           &           &           &           &           &           & \like{6}         \\
\textbf{Spatial data}                  &           &           &           &           &           &           & \like{6} \\
Time                          & \like{6} & \like{6} & \like{6} & \like{6} & \like{6} & \like{6} &  \\
Environmental sensors & \like{6} & \like{6} & \like{6} &           &           &           &           \\
\textbf{Near body temperature}         &           & \like{6} & \like{6} & \like{6} &           &           & \like{6} \\
\textbf{Heart rate}                    &           & \like{6} & \like{6} & \like{6} & \like{6} &           & \like{6} \\
Room                          &           &           & \like{6} & \like{6} & \like{6} & \like{6} &           \\
Preference history            &           &           & \like{6} & \like{6} & \like{6} & \like{6} &           \\ \midrule
\textbf{Model accuracy}                & 0.58      & 0.65      & 0.7       & 0.72      & 0.68      & 0.64      & {\textbf{0.86}}     \\ \bottomrule
\end{tabular}
\end{table*}

For the proposed model using Build2Vec, we used four features as inputs in the model to compare to the baselines. These features were selected in such a way that they are linked to the geolocation of the participant during the thermal comfort preference feedback point. Thus, these features become attributes of both the location and the user. Additionally, in the modeling process, participants were clustered into groups of similar preferences in a process that aligns with the baseline \cite{Jayathissa2020-pv}. These features used in the proposed model are:
\begin{enumerate}
    \item \textbf{User location} - location of the occupant that is merged with the graph structure of the building, then embedding vector.
    \item \textbf{Spatial data} -  data that are converted into a graph representation, then converted into an embedding vector. 
    \item \textbf{Near-body temperature} -  the most representative temperature at the exact location when the thermal preference feedback was given.
    \item \textbf{Heart rate} - used as a proxy for the level of activity (metabolic rate). 
\end{enumerate}


The results of the implementation of the Build2Vec method showed a training accuracy of 0.92, validation accuracy of 0.93, and testing accuracy of 0.861 using only the Build2Vec embedding vector, heart rate, and near body temperature as features. It can be noticed that the proposed method outperforms the baseline, which confirms the validity of the hypothesis in this case that the spatial proximity to different spatial objects has improved prediction capabilities for thermal perception. Also, the spatial similarity metric (Build2Vec) could be used to compensate for the lack of sensors resources. This model can be used as a novel method for automatic thermal zoning for simulation.

\section{Discussion}
\label{sec:discussion}
The implementation of this case study opens new directions related to using intensive longitudinal data from groups of occupants in ways that can improve the characterization of spaces. This section discusses the impact of the results on the development and use of BIM models in the operations of buildings, utilizing the subjective feedback of human occupants in better ways, and the use of such information for post-occupancy evaluation and building controls. In addition, limitations to the generalizability of this case study implementation results are discussed to provide a foundation for future work.

\subsection{Impact of digital twins on thermal comfort prediction}
\label{sec:impact}
Utilizing the spatial context for thermal comfort prediction can influence the way BIM models are used and the nudging the use of these models in the new \emph{digital twin} context. BIM models are generally developed in the design phase of the building life cycle and are only beginning to be leveraged in the operations phase. Attaching spatially and temporally dynamic data to BIM effectively brings these spatial representations into the digital twin paradigm. This shift is happening in numerous domains through the real-time convergence of data sets that are conventionally siloed~\cite{Miller2021-ql}.

\subsection{Using humans-as-sensors in buildings}
\label{sec:IoT_integration}
This study illustrates that the spatial context and human subjective feedback can supplement the deployment of IoT IEQ sensors. It is impractical for sensors to be deployed at such a high resolution to measure every micro-climate situation in a building. This methodology leverages intensive longitudinal data collection from several occupants to \emph{use humans} and their ability to sense and interpret far more information than any sensor kit. Deployment of such a methodology using a smartwatch makes this collection feasible as the compliance rate for micro-ecological momentary assessments is high~\cite{Intille2016-xb}. 

\subsection{Possibilities for design-phase feedback in post-occupancy evaluation}
\label{sec:poe}
Techniques like the one presented combined with intensive longitudinal data could supplement evaluating which design features result in the intended result. Design options such as glazing ratio and positioning and heating or cooling terminal unit or diffuser location could be assessed by meshing the subjective feedback data with the spatial context. Post-occupancy evaluations are designed to capture such information but lack the data from each occupant for the diversity of spaces that could give insight into specific design features~\cite{Li2018-yz}.

\subsection{Possibilities for occupant-centric controls integration}
The framework outlined could produce predictions integrated into building controls and automation systems to optimize building controls operations. The progression of occupant-centric controls welcomes using signals from alternative data sources, and the spatial features developed in this work might be influential in controls algorithms~\cite{OBrien2020-ns,Gunay2021-ly}. Several existing studies investigate the use of personal comfort models for automation and controls, and Build2Vec could be tested in these contexts~\cite{Li2017-fh, Park2018-cu}.



\subsection{Spatial recommendation}
Given the accuracy of the current model, this model can be used as a spatial recommendation engine. For example, if the user's heart rate and near body temperature data are known, this model can recommend specific cells that are likely to be comfortable. This method also can be informative to the real estate managers and facility managers to allocate furniture and employees based on the employees' satisfaction preferences. The process of collecting and predicting subjective occupant data for indoor environmental quality goes beyond just thermal comfort. A prominent study investigated lighting, noise, and general spatial satisfaction as well~\cite{Kim2013-gs}. 

\subsection{Spatial clustering}
Clustering of spaces or part of spaces is yet another potential application of this methodology. For example, locations near the window and ceiling fan can be clustered in the same spatial groups. Using the Build2Vec spatial clustering method, spaces, parts of spaces, floors, and whole buildings can be clustered in different scales based on their similarity. This method can be informative in terms of energy benchmarking of buildings, among other applications. 

\subsection{Limitations of results from this case study implementation}
In the context of any machine learning-based methodology, one should always understand the limitations of generalizing the results beyond the experiment, primarily when a single case study is used to showcase a method. In this paper, the framework has been applied to a single building in Singapore. In order to draw broader conclusions, the technique should be repeated and expanded to case studies and occupant test sizes that are much larger and more diverse.

\section{Conclusion}
\label{sec:conclusion}
This study explores the relation of occupants' location in a building to their indoor environmental satisfaction perception. The occupants' location defines their proximity to different building elements (spatial proximity), such as windows, fans, diffusers, or heat sources, which significantly impact their satisfaction perception. To address the spatial proximity issue, an end-to-end spatial-temporal prediction model called Build2Vec has been developed. This model automatically captures the location of occupants using an indoor positioning system. This indoor positioning system is coupled with occupants' indoor satisfaction feedback, which is collected via ecological momentary assessment method (using smartwatches) over a time period. Both the location data and the occupants' feedback are combined with indoor environmental data from the sensors and the spatial data from the BIM data to generate the final resultant model. The resulting model is then converted into a temporal graph network to formulate the thermal preference prediction model. Afterward, a classification model is used to predict an occupant's thermal preference based on their current and historical location. The result shows that Build2Vec outperforms the baseline model noticeably. Reproducible code and data for this publication can be found in the following Github repository: \url{https://github.com/buds-lab/build2vec-thermal-comfort}.



\section*{CRediT Author statement}
\textbf{MA:} Conceptualization, Methodology, Validation, Formal analysis, Writing - Original Draft, and Visualization, Data Curation, and Investigation. \textbf{AC:} Writing - Review \& Editing. \textbf{CM:} Conceptualization, Methodology, Investigation, Resources, Data Curation, Writing - Review \& Editing, Supervision, Project administration, and Funding acquisition.

\section*{Funding}
Singapore Ministry of Education (MOE) Tier 1 Grants provided support for the development and implementation of this research under the projects Ecological Momentary Assessment (EMA) for Built Environment Research (R296000210114) and The Internet-of-Buildings (IoB) Platform – Visual Analytics for AI Technologies towards a Well and Green Built Environment (R296000214114).

\section*{Acknowledgements}
The authors would like to acknowledge the team behind the dataset collection and processing Prageeth Jayathissa, Matias Quintana, Yi Ting Teo, Yun Xuan Chua, and Charlene Tan. Also, the authors would like to thank NUS for facilitating the deployment of sensors and BLE beacons across its buildings. This work contributes to the IEA EBC - Annex 79 - Occupant-Centric Building Design and Operation.

\bibliographystyle{model1-num-names}
\bibliography{references}

\end{document}